\documentclass{article}

    \PassOptionsToPackage{numbers, compress}{natbib}

    \usepackage[preprint]{nips_2018}

\usepackage[utf8]{inputenc} 
\usepackage[T1]{fontenc}    
\usepackage{hyperref}       
\usepackage{url}            
\usepackage{booktabs}       
\usepackage{amsfonts}       
\usepackage{nicefrac}       
\usepackage{microtype}      
\setcitestyle{square}

\usepackage{times}
\usepackage{epsfig}
\usepackage{graphicx}
\usepackage{amsmath}
\usepackage{amssymb}
\usepackage{makecell}
\usepackage{threeparttable}
\usepackage{caption}

\usepackage{mathrsfs}
\usepackage{booktabs}
\usepackage{threeparttable}
\usepackage{multirow}
\usepackage{amsfonts,amssymb}
\usepackage{setspace}
\usepackage{float}
\usepackage{subcaption}

\begin{document}

\title{LVIS Challenge Track Technical Report 1st Place Solution: Distribution Balanced and  Boundary Refinement for Large Vocabulary Instance Segmentation}

\author{
  WeiFu Fu\thanks{indicates equal contribution.} \ ,  \  CongChong Nie\footnotemark[1] \ ,\ Ting Sun, \ Jun Liu, \ TianLiang Zhang, \ and Yong Liu\\\\
  Tencent YouTu Lab\\
  \texttt{ryanwfu@tencent.com, nickncc2122@gmail.com, yolandasun@tencent.com,} \\
  \texttt{junsenselee@gmail.com, linuszhang@tencent.com, choasliu@tencent.com}
}

\maketitle

\begin{abstract}
  This report introduces the technical details of the team FuXi-Fresher for LVIS Challenge 2021. Our method focuses on the problem in following two aspects: the long-tail distribution and the segmentation quality of mask and boundary. Based on the advanced HTC instance segmentation algorithm, we connect transformer backbone(Swin-L) through composite connections inspired by CBNetv2 to enhance the baseline results. To alleviate the problem of long-tail distribution, we design a Distribution Balanced method which includes dataset balanced and loss function balaced modules. Further, we use a Mask and Boundary Refinement method composed with mask scoring and refine-mask algorithms to improve the segmentation quality. In addition, we are pleasantly surprised to find that early stopping combined with EMA method can achieve a great improvement. Finally, by using multi-scale testing and increasing the upper limit of the number of objects detected per image, we achieved more than 45.4\% boundary AP on the val set of LVIS Challenge 2021. On the test data of LVIS Challenge 2021, we rank 1st and achieve 48.1\% AP. Notably, our APr 47.5\% is very closed to the APf 48.0\%.
\end{abstract}

\section{Introduction}

LVIS \cite{lvis} is a new benchmark dataset for large vocabulary instance segmentation. It contains about 2 million high-quality instance segmentation annotations in more than 1,000 categories. Compared with the COCO \cite{coco} dataset, the LVIS dataset has a long-tail distribution which is more similar to the real world. According to the statistics, the rare categories only accounts for about 1/3 of the total number of categories and 0.41\% of the total number of instances. Thus, the unbalanced data increases the difficulty of training which is the first challenge in LVIS workshop. In addition, more and more attention has been paid to more precise segmentation result. Therefore, the boundary AP evaluation metric \cite{boundaryap} is proposed and puts forward higher requirements on the mask prediction quality. However, many algorithms fail to well segment the boundary of mask. It is the other challenge in LVIS workshop. 

To solve these challenges, our solution proposed two algorithmic modules: (1) Distribution Balanced Algorithm: including RFS, balanced copy-paste and balanced Mosaic augmentation and Seesaw loss function to deal with long-tail distribution; (2) Mask and Boundary Refinement Algorithm: including re-scoring the mask prediction and refinemask methods to optimize the quality of mask and boundary. Overall, based on effective data augmentation strategies, an optimized loss function and a stronger backbone with the two-stage HTC \cite{htc} model, we improve the performance on long-tail distribution problem. Further, refinemask \cite{refinemask} gradually integrate fine-grained features and refine the mask, and mask scoring re-scoring the mask prediction. The details will be described in Section 2.

In addition, we utilized early stopping combined with EMA method in learning based on our the observation of the experiment. The method has a great improvement and is robust. The details will be described in Section 3.2.

Overall, On the test data of LVIS Challenge 2021, we achieve 48.1\% AP. Notably, our APr 47.5\% is very closed to the APf 48.0\%.

\section{Our Solution}

\subsection{Strong Baseline}
\textbf{(1) Hybrid Task Cascade (HTC):} HTC is a classic yet powerful cascade architecture for instance segmentation task. It interweaves detection and segmentation branches for a joint multi-stage processing and progressively  learns more discriminative features in each stage. In our solution, we removed the semantic head since it requires extra stuff segmentation annotations.

\textbf{(2) Swin Transformer \cite{swin} and CBNetV2 \cite{cbnetv2} Backbone:} Vision transformer has made a great progress in many visual tasks recently. Thus, we use Swin Transformer as our backbone. Swin Transformer proposed an efficient window attention module in a hierarchical feature architecture. It has a linear computational complexity with respect to the input image size. In this work, we use Swin-L network as our basic backbone which is pretained by Imagenet-22k dataset. Furthermore, inspired by CBNetv2 algorithm, we groups two identical Swin-L networks through composite connections as the final backbone to enhance the performance.

\subsection{Distribution Balanced}

\textbf{(1) Repeat factor sampling (RFS) \cite{lvis}:} Taking into account the serious imbalance of the LVIS dataset, we use RFS to increase the sampling rate of the images containing the tail categories.

\textbf{(2) Balanced Copy-Paste:} Considering that RFS is an image-level re-sampling technology, we further adopt copy-paste \cite{copypaste} data augmentation method, which can be considered as an instance-level re-sampling technology. It can increase the sampling probability of the tail categories instances, and can avoid the mixing of the head categories instances caused by RFS.

\textbf{(3) Balanced Mosaic:} The balanced mosaic \cite{mosaic} strategy randomly samples among four training images for stitching based on the RFS re-sampling dataset. It can significantly increase the number of instances in each image and enrich the background. We note that the scale of instance size will be reduced relatively in the generated image which will result in some distribution difference between the train set and the test set. Therefore, We use a probability of 0.5 to randomly choose training images to do balanced mosaic for avoiding distribution shift.

\subsection{Loss Function}

\textbf{Seesaw Loss:} Seesaw Loss \cite{seesaw} uses mitigation factor and compensation factor to dynamically suppress excessive negative sample gradients on the tail category, while supplementing the penalty for misclassified samples to avoid false positives.

\subsection{Mask and Boundary refinement}
Although our strong baseline has achieved a great improvement, the instance mask generated by the HTC model is still very coarse due to the pooling process and the 28x28 masks representation size. Therefore, we use refinemask to replace the mask head of the last stage of HTC to obtain a higher quality mask prediction. At the same time, we also used mask scoring to re-scoring the final mask to further improve the mask metric.

\textbf{(1) Mask Scoring:} Directly using classification confidence as the mask score will result in misregistration of the mask score and the mask quality. Mask scoring \cite{maskscoring} can automatically learn the mask quality instead of relying on the classification confidence of the bounding box.

\textbf{(2) RefineMask:} The refinemask \cite{refinemask} gradually integrates fine-grained features and designs boundary supervision to obtain high-quality masks. Referring to the refinemask of the maskRCNN version, we initially replaced the three mask heads of HTC with refinemask heads. However, this will result in a significant increase in GPU memory and training time. In order to balance the training time and accuracy, we only replace the last mask head with a refinemask head. In addition, we set the loss weight of bbox head to 2.0 to balance the weight between bbox head and mask head.

\section{Experiments}

\subsection{Implementation Details}
In our experiments, we use HTC without semantic branch as our baseline. The batch size is set to 16. The learning rate is set to 0.0001 and 0.00005 corresponding to single swin-L backbone and CBNetV2 backbone. We train all models for total 20 epochs by default, with learning rate divided by 10 at the 16th and 19th epoch. All models are initialized with ImageNet-22k pre-trained model.

\subsection{Experimental Results}

\subsubsection{Distribution Balanced and Strong Baseline}
We use RFS and seesaw Loss to deal with the long tail problem from the perspective of data re-sampling and loss function. Note that, we directly adopt Swin-L as the basic backbone because some methods are effective on a simple backbone but do not work on a stronger backbone. The basic experimental results in Table \ref{table1} achieved 39.1\% mask AP (\textbf{named Weak Baseline}), which directly surpasses the last year's third place. Subsequently, experiments with balanced copy-paste and balanced mosaic achieves mask AP improvements of 0.7\% and 1.0\%, respectively. In addition, the improvement of $AP_r$ is more significant, achieving an improvement of 3.9\% and 1.5\% respectively. After adopting a stronger CBNetV2 backbone, mask AP has been further improved to 43.1\% (\textbf{named Strong Baseline}), and $AP_r$, $AP_c$, and $AP_f$ have also been greatly improved.

\begin{table}[h]
\renewcommand\arraystretch{1.5}
\begin{center}
\setlength{\tabcolsep}{6pt}
\begin{tabular}{ccccccccccc}
\toprule[1.2pt]
RFS & SL & Swin-L & B-C & B-M & CBV2  & mask AP(\%)  & $AP_r(\%)$  & $AP_c(\%)$  & $AP_f(\%)$  \\
\hline
                     \checkmark  & \checkmark  & \checkmark  &             &            &             & 39.1  & 27.9  & 39.2  & 43.7 \\
                     \checkmark  & \checkmark  & \checkmark  & \checkmark  &            &             & 39.8  & 31.8  & 40.2 & 42.8  \\
                     \checkmark  & \checkmark  & \checkmark  & \checkmark  & \checkmark &             & 40.8	&33.3	&40.5	&44.3\\
      \checkmark  & \checkmark  & \checkmark  & \checkmark  & \checkmark & \checkmark  & 43.1	&34.3	&43.1	&47.0\\
\bottomrule[1.2pt]
\end{tabular}
\setlength{\abovecaptionskip}{10pt}%
\setlength{\belowcaptionskip}{0pt}%
\caption{Ablation studies on LVIS v1.0 val set. Models are HTC w/o semantic branch. RFS: Repeat Factor Sampling. SL: Seesaw Loss. B-C: Balanced Copypaste. B-M: Balanced Mosaic. CBV2: CBNetV2. We also used multi-scale training, random flip, crop and color jitter in all experiments. It is worth noting that the experiments using the balanced mosaic adopt the short-side zoom range of (640, 1400), and the other experiments use the zoom range of (400, 1400), which is to avoid generating too many extremely small objects.}
\label{table1}
\end{center}
\end{table}

\subsubsection{Ablation of Refinemask}

We only conduct the refinemask ablation experiment on Week Baseline due to time constraints. As shown in Table \ref{Refinemask}, refinemask can lead to +1.1\% in mask AP. Meanwhile, the metrics of $AP_r$, $AP_c$ and $AP_f$ also has an improvement. Thus, we infer that refinemask can also improve the performance on Strong Baseline. It’s worth noting that we only replace the last mask head with a refinemask head in HTC.

\begin{table}[h]
\renewcommand\arraystretch{1.5}
\begin{center}
\setlength{\tabcolsep}{5.6pt}
\begin{tabular}{cccccccc}
\toprule[1.2pt]
Method	&epoch		&mask AP  & $AP_r(\%)$  & $AP_c(\%)$  & $AP_f(\%)$  \\
\hline
Weak Baseline	&20		&39.1	&27.9	&39.2	&43.7 \\
+refinemask	&20		&40.2	&29.1	&40.3	&44.9\\
\bottomrule[1.2pt]
\end{tabular}
\setlength{\abovecaptionskip}{10pt}%
\setlength{\belowcaptionskip}{0pt}%
\caption{Refinemask ablation studies on LVIS v1.0 val set based on a Weak Baseline.}
\label{Refinemask}
\end{center}
\vspace{-5mm}
\end{table}

\subsubsection{Early Stopping and EMA}
\begin{figure*}[h]
\centering
\includegraphics[width=14cm]{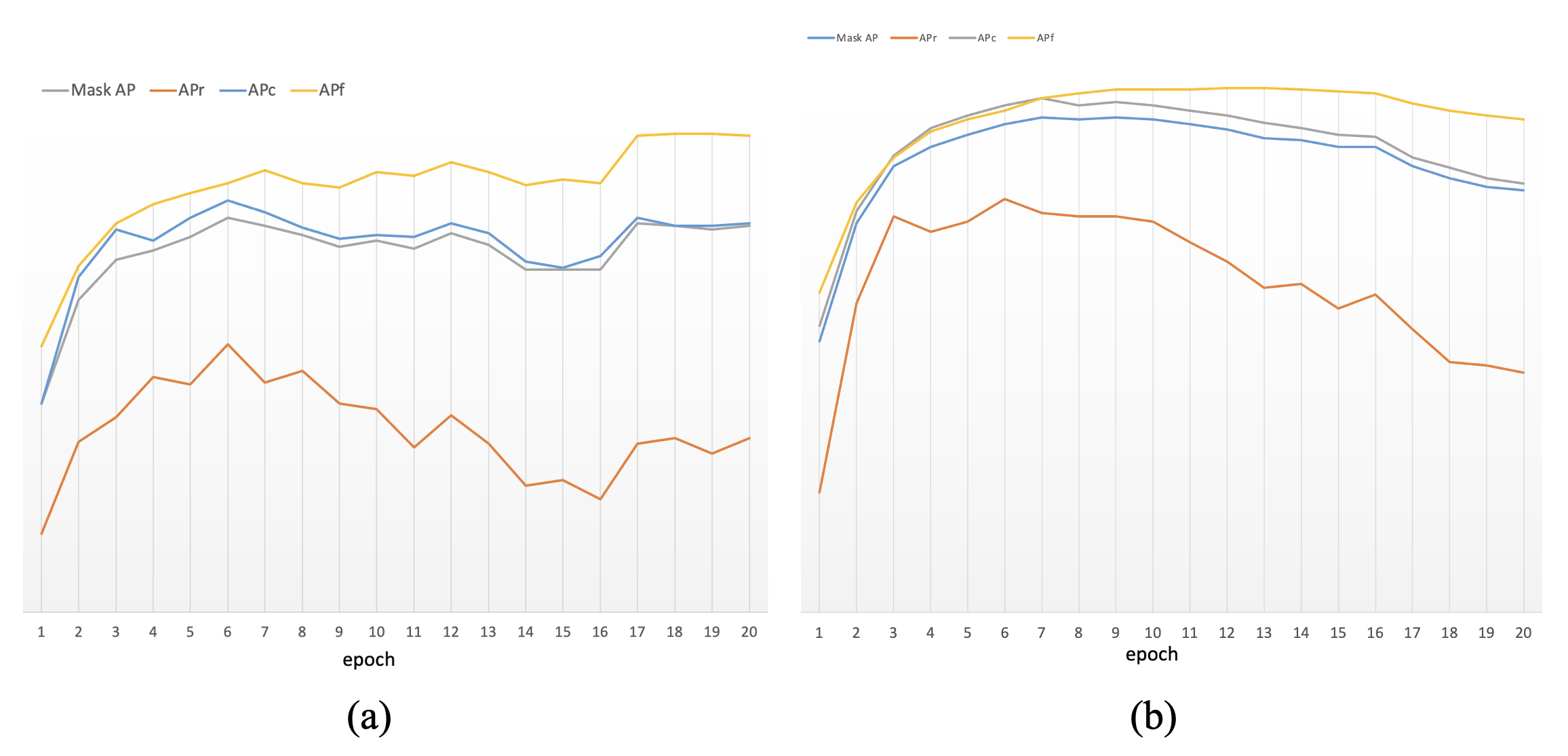}
\caption{AP curve on the LVIS validation set. (a) AP curve of strong baseline model in row 1 of Table \ref{table2}. (b) AP curve of model in row 2 of Table \ref{table2}.}
\label{ema}
\end{figure*}

As shown in Figure \ref{ema} (a), we visualize the AP curve of the strong baseline on LVIS validation set.  It can be seen that the curves of $AP_r$ and $AP_f$ have a large contrast, reaching their peaks at 6th epoch and 20th epoch, respectively. However, the mask AP of 6th epoch and 20th epoch are very similar. We think this is caused by the long tail distribution problem. 

In addition, considering the large fluctuation in the performance of the strong baseline model, we also adopt the Exponential Moving Average (EMA) training strategy to smooth the model weights. As shown in Figure \ref{ema} (b), we visualize the AP curve of the model in row 2 of Table \ref{table2}. It can be seen that the mask AP, $AP_{r}$ and $AP_{c}$ at 6th epoch both obtain a remarkable increase compared with 20th epoch. Above all, the gap between $AP_{r}$, $AP_{c}$ and $AP_{f}$ is also significantly resolved. Therefore, we finally adopted the early stopping and EMA strategy in subsequent models.

\subsubsection{Mask and Boundary refinement}

On the basis of strong baseline, we report the experimental results of refinemask and mask scoring in Table \ref{table2}. In particular, we use EMA, early stopping and FP32 in row2 and row3 of Table \ref{table2} to further improve accuracy. It can be seen that the combination of EMA, early stopping, FP32 and refinemask can improves the mask AP by 5\%. After further integrating mask scoring, mask AP obtains a higher result of 49.2 mask AP on the val set.

\begin{table}[h]
\renewcommand\arraystretch{1.5}
\begin{center}
\setlength{\tabcolsep}{3.5pt}
\begin{tabular}{ccc|cccccc}
\toprule[1.2pt]
EMA & early stop&FP32	&refinemask	&mask scoring	&Mask AP(\%)	& $AP_r(\%)$  & $AP_c(\%)$  & $AP_f(\%)$  \\
\hline
& &	&	&	&43.1	&34.3	&43.1	&47.0  \\
\checkmark&\checkmark&\checkmark	&\checkmark	&	&48.1	&43.8	&49.2	&48.9.  \\
\checkmark&\checkmark&\checkmark	&\checkmark	&\checkmark	&49.2	&45.4	&50.0	&50.0.  \\
\bottomrule[1.2pt]
\end{tabular}
\setlength{\abovecaptionskip}{10pt}%
\setlength{\belowcaptionskip}{0pt}%
\caption{Mask scoring and refinemask ablation studies on LVIS v1.0 val set. Both mask scoring and refinemask head are only added to the last stage of HTC.}
\label{table2}
\end{center}
\vspace{-5mm}
\end{table}

\section{Final Results}

\subsection{Test Time Augment (TTA)}
As shown in Table \ref{tta}, we use standard multi-scale testing, which adopts multiple test resolutions ((1600, 1000), (1600, 1400), (1800, 1200), (1800, 1600)) and flip strategy.

\begin{table}[h]
\renewcommand\arraystretch{1.5}
\begin{center}
\setlength{\tabcolsep}{10pt}
\begin{tabular}{ccccccc}
\toprule[1.2pt]
TTA	& Mask AP(\%) & Boundary AP(\%)	& $AP_r$(\%)  & $AP_c$(\%)  & $AP_f$(\%)  \\
\hline
        & 49.2    &44.1	&40.4	&45.3	&44.4 \\
\checkmark	&50.4 & 45.4	&41.1	&46.5	&45.9 \\
\bottomrule[1.2pt]
\end{tabular}
\setlength{\abovecaptionskip}{10pt}%
\setlength{\belowcaptionskip}{0pt}%
\caption{Test time augmentation ablation studies on LVIS v1.0 val set.}
\label{tta}
\end{center}
\vspace{-5mm}
\end{table}

\subsection{More Detections Per Image}
In this year’s competition rules, there is no limit on the number of detections per image. As shown in Table \ref{max}, we increase the number of detections per images to 1000 finally.

\begin{table}[H]
\renewcommand\arraystretch{1.2}
\begin{center}
\setlength{\tabcolsep}{10pt}
\begin{tabular}{ccccc}
\toprule[1.2pt]
max\_per\_img	& Boundary AP(\%)	& $AP_r$(\%)  & $AP_c$(\%)  & $AP_f$(\%)  \\
\hline
300		&44.1	&40.4	&45.3	&44.4 \\
500		&44.9	&42.6	&45.9	&44.8 \\
800	    &45.2	&43.4	&46.2	&44.9 \\
1000	&45.3	&43.5	&46.3	&44.9 \\
\bottomrule[1.2pt]
\end{tabular}
\setlength{\abovecaptionskip}{10pt}%
\setlength{\belowcaptionskip}{0pt}%
\caption{More detections per image ablation studies on LVIS v1.0 val set.}
\label{max}
\end{center}
\vspace{-5mm}
\end{table}

\subsection{Test Challenge Result}

Our final model applies TTA and increases the number of objects detected per image to 1000, which will exceed 45.4\% boundary AP on LVIS v1.0 val set. Due to time constraints, we did not test its results on the LVIS v1.0 val set, but on LVIS v1.0 Test Challenge, we used both in the final submission. Our final result is shown in Table \ref{final}. It can be seen that our result has almost equal results on $AP_r$, $AP_c$ and $AP_f$, which confirms the effectiveness of our solution on the long-tail distribution problem.

\begin{table}[H]
\renewcommand\arraystretch{1.5}
\vspace{5mm}
\begin{center}
\setlength{\tabcolsep}{8pt}
\begin{tabular}{ccccccc}
\toprule[1.2pt]
Set & TTA	& max\_per\_img & $AP$†(\%)	& $AP_r$†(\%)  & $AP_c$†(\%)  & $AP_f$†(\%)  \\
\hline
Val & 	\checkmark  &300 	& 45.4	&41.1	&46.5	&45.9 \\
Val & 	  &1000 	&45.3	&43.5	&46.3	&44.9\\
Test challenge & \checkmark & 1000	&48.1	&47.5	&48.5	&48.0 \\
\bottomrule[1.2pt]
\end{tabular}
\setlength{\abovecaptionskip}{10pt}%
\setlength{\belowcaptionskip}{0pt}%
\caption{Final Result on LVIS v1.0 Test Challenge. AP†: refers to the "boundary, fixed AP \cite{dave2021evaluating}" used in the LVIS challenge 2021.}
\label{final}
\end{center}
\vspace{-5mm}
\end{table}


{\small
\bibliographystyle{ieeetr}
\bibliography{ref}
}

\end{document}